\documentclass[sigconf]{acmart} %% camera-ready：显示作者、致谢；无审稿行号
\graphicspath{{figures/}}
\usepackage{enumitem}
\copyrightyear{2026}
\acmYear{2026}
\setcopyright{cc}
\setcctype{by}
\acmConference[ICMR '26]{International Conference on Multimedia Retrieval}{June 16--19, 2026}{Amsterdam, Netherlands}
\acmBooktitle{International Conference on Multimedia Retrieval (ICMR '26), June 16--19, 2026, Amsterdam, Netherlands}
\acmDOI{10.1145/3805622.3810761}
\acmISBN{979-8-4007-2617-0/2026/06}

\begin{document}
%% 1. 标题
\title{MOC-3D: Manifold-Order Consistency for Text-to-3D Generation}
%% 2. 作者信息（camera-ready）

%%
%% The "author" command and its associated commands are used to define
%% the authors and their affiliations.
%% Of note is the shared affiliation of the first two authors, and the
%% "authornote" and "authornotemark" commands
%% used to denote shared contribution to the research.
\author{Chenyang Fan}
\authornote{Chenyang Fan and Junshi Cheng contributed equally to this work.}
\email{2024210516029@stu.cqnu.edu.cn}
\orcid{0009-0004-8232-8217}
\affiliation{%
  \institution{Chongqing Normal University}
  \city{Chongqing}
  \country{China}
}

\author{Junshi Cheng}
\authornotemark[1]
\email{2023210516027@stu.cqnu.edu.cn}
\orcid{0009-0000-3422-2865}
\affiliation{%
  \institution{Chongqing Normal University}
  \city{Chongqing}
  \country{China}
}

\author{Wen Yang}
\authornote{Corresponding author.}
\email{yangwen@cqnu.edu.cn}
\orcid{0000-0002-0683-7958}
\affiliation{%
  \institution{Chongqing Normal University}
  \city{Chongqing}
  \country{China}
}

\author{Zihong Li}
\email{LiZiHong20001026@126.com}
\orcid{0009-0008-1547-861X}
\affiliation{%
  \institution{Chongqing Normal University}
  \city{Chongqing}
  \country{China}
}

\author{Wenfeng Zhang}
\authornotemark[2]
\email{itzhangwf@cqnu.edu.cn}
\orcid{0000-0001-7459-2510}
\affiliation{%
  \institution{Chongqing Normal University}
  \city{Chongqing}
  \country{China}
}

\author{Wei Hu}
\email{wei369.workstation@cqnu.edu.cn}
\orcid{0000-0002-4637-8995}
\affiliation{%
  \institution{Chongqing Normal University}
  \city{Chongqing}
  \country{China}
}

\author{Yi Zhang}
\email{zhangyii@cqnu.edu.cn}
\orcid{0009-0004-4633-1363}
\affiliation{%
  \institution{Chongqing Normal University}
  \city{Chongqing}
  \country{China}
}

\author{Pan Zeng}
\email{zengpan@cqnu.edu.cn}
\orcid{0000-0002-6674-9749}
\affiliation{%
  \institution{Chongqing Normal University}
  \city{Chongqing}
  \country{China}
}

\renewcommand{\shortauthors}{Fan, Cheng, Yang, et al.}

%% 3. 摘要
\begin{abstract}
With the burgeoning development of fields such as the Metaverse, Virtual Reality (VR), and Digital Twins, text-to-3D generation has emerged as a research hotspot in both academia and industry. Currently, optimization methods based on Score Distillation Sampling (SDS) utilizing 2D diffusion priors have become the mainstream technological paradigm in this field. However, due to the view bias of 2D priors and the mode-seeking ambiguity combined with gradient noise induced by high Classifier-Free Guidance (CFG), these methods still suffer from macro-topological inconsistency (e.g., the Janus problem) and micro-geometric discontinuity.
To address these challenges, we propose MOC-3D, a text-to-3D generation method based on geometric manifold and semantic view-order consistency. Built upon the ScaleDreamer framework, our method incorporates a Semantic View-Order Constraint Module and a Manifold-based Feature Continuity Module. The former aims to rectify macro-topological inconsistency, while the latter focuses on eliminating micro-geometric discontinuity. Specifically, the Semantic View-Order Constraint Module leverages the prior knowledge of CLIP to impose a Monotonicity Rank Constraint on semantic score representations across different views, thereby providing effective guidance for the global topological structure of 3D objects. Meanwhile, the Manifold-based Feature Continuity Module employs the Riemannian Metric on the Symmetric Positive Definite (SPD) manifold. By measuring the distance of feature statistical distributions in the Riemannian space, it promotes the smooth evolution and continuity of micro-textures across multi-views in a statistical sense. Under the macro-micro synergistic optimization of these two modules, our model can simultaneously improve macro-structural consistency and micro-detail continuity.
Experimental results demonstrate that compared with mainstream methods, our approach achieves significant advantages in terms of Semantic Consistency (CLIP Score) and Perceptual Quality (LPIPS). Furthermore, ablation studies verify the independent contributions and complementary effectiveness of the Semantic View-Order Constraint Module in rectifying macro-topological inconsistency and the Manifold-based Feature Continuity Module in eliminating micro-geometric discontinuity.
\end{abstract}

%% 4. CCS（来自 dl.acm.org/ccs/ccs.cfm “View CCS TeX Code”）

%% 5. 关键词
\maketitle
%% ================= 正文区域 =================
\section{Introduction}
Text-guided 3D generation technology is reshaping traditional 3D digital content creation modes~\cite{hu2025turbo3d,xiang2025structured,liang2024luciddreamer,qiu2024richdreamer}, providing efficient productivity tools for fields such as Metaverse construction, film and game development, and Digital Twins. Aiming to automatically generate high-fidelity 3D digital assets via natural language descriptions, this technology holds immense academic research value and broad application prospects.
Currently, technological paradigms in this field are primarily categorized into two types: feed-forward generation methods relying on large-scale 3D datasets (e.g., Point-E~\cite{nichol2022point}, Shap-E~\cite{jun2023shape}), and optimization-based methods leveraging pre-trained 2D text-to-image diffusion models. Due to the high cost and scarcity of high-quality 3D data, the former often faces challenges in texture details and generalizability. In contrast, optimization-based methods based on 2D diffusion priors have gradually become the mainstream research direction. Unlike recent 3D Gaussian Splatting methods that prioritize speed but often lack surface connectivity, optimization-based approaches ensure topology-ready asset generation, offering powerful zero-shot capabilities without relying on 3D data.
Within this mainstream direction, the Score Distillation Sampling (SDS) framework pioneered by DreamFusion~\cite{poole2023dreamfusion} has achieved remarkable progress. However, since the guiding 2D diffusion models inherently lack 3D spatial awareness, they optimize each viewpoint independently without explicit cross-view spatial correlation, and the fitting process lacks statistical alignment mechanisms for high-dimensional features of regular details, these methods are prone to generating severe structural topological anomalies (i.e., the ``Janus problem'' such as multi-faced portraits or multi-headed animals) as well as micro-geometric noise and misalignments.
To mitigate these challenges, subsequent works have made significant improvements from various perspectives. Magic3D~\cite{lin2023magic3d} and Fantasia3D~\cite{chen2023fantasia3d} introduced a ``coarse-to-fine'' two-stage optimization strategy to sequentially fit geometric shapes and refine surface textures. While these methods successfully enhance visual quality, ensuring global macro-topological consistency remains an open challenge. Similarly, methods like MVDream~\cite{shi2024mvdream} incorporated multi-view diffusion models fine-tuned on 3D datasets to provide consistency guidance. Although this approach effectively alleviates the Janus problem for common objects, its generalizability can be influenced by the scale of the 3D training data, particularly when handling complex or abstract text prompts that deviate from the training distribution. Meanwhile, ProlificDreamer~\cite{wang2023prolificdreamer} proposed Variational Score Distillation (VSD), which elegantly addresses the texture over-saturation issue via 3D parameter distribution modeling. Despite its success in texture generation, current methods struggle to achieve rigorous global topological consistency and eliminate micro-level geometric discontinuities. Furthermore, their high computational overhead underscores the need for more efficient optimization strategies. In summary, balancing ``macro-semantic constraints'' and ``micro-texture statistical constraints'' to achieve physically consistent geometric structures alongside high-fidelity surface details within a single framework remains a key research objective.
To further address the challenges of macro-topological inconsistency and micro-geometric discontinuity, we propose MOC-3D, a text-to-3D generation method based on geometric manifold and semantic view-order consistency. For macro-consistency, we design a Semantic View-Order Constraint Module, which introduces score evaluation based on CLIP priors and incorporates a semantically consistent Monotonicity Rank Constraint, aiming to better guide the global topological structure. For micro-continuity, we introduce a Manifold-based Feature Continuity Module. By measuring the statistical distribution distance of multi-view features based on the Symmetric Positive Definite (SPD) manifold, this module seeks to align the high-dimensional regularity of surface textures.
Our main contributions are summarized as follows:
\begin{itemize}
    \item We present MOC-3D, a novel text-to-3D generation framework employing dual constraints of manifold and semantics. By synergizing macro-structural guidance with micro-statistical alignment, it achieves high-fidelity detail generation while maintaining semantic consistency, surpassing existing methods in quality.
    \item To mitigate macro-topological inconsistencies, we design a Semantic View-Order Constraint Module. By leveraging CLIP priors to enforce a Monotonicity Rank Constraint, this module effectively addresses the view bias inherent in 2D priors and rectifies structural anomalies.
    \item To address micro-geometric discontinuities and high-frequency artifacts, we introduce a Manifold-based Feature Continuity Module. By mapping multi-view features into the Riemannian manifold space and aligning their statistical distributions, we effectively eliminate micro-geometric discontinuities and high-frequency artifacts induced by gradient noise.
\end{itemize}
\section{Related Work}
\textbf{Text-to-3D Generation.} With the burgeoning development of large-scale cross-modal pre-training models, text-to-3D generation has witnessed significant progress in recent years. Broadly speaking, technical routes in this field can be categorized into two main streams: direct feed-forward inference methods and optimization-based generation methods.
Feed-forward methods typically prioritize inference speed. While early attempts offered rapid generation, they were often constrained by limited texture quality and generalization capabilities. To alleviate these bottlenecks, several Large Reconstruction Models (LRMs)~\cite{hong2024lrm} have significantly advanced generation performance. For instance, LGM~\cite{tang2024lgm} leverages 3DGS to achieve high-resolution generation within seconds; Turbo3D~\cite{hu2025turbo3d} introduces dual-teacher distillation to balance sub-second speed with quality; and TRELLIS~\cite{xiang2025structured} employs structured latent variables to effectively handle complex topologies. Despite their exceptional inference speeds, the generalization capability of such methods remains heavily reliant on the coverage of 3D training data, and fine-grained semantic control is often limited. In contrast, optimization-based methods leverage pre-trained 2D diffusion priors to guide iterative optimization. Obviating the need for 3D training data, these methods achieve zero-shot generation and have emerged as the dominant paradigm due to their superior capability in generating high-quality and stylized assets.
As a pioneering work in optimization-based generation, DreamFusion~\cite{poole2023dreamfusion} introduced Score Distillation Sampling (SDS), innovatively utilizing a pre-trained 2D text-to-image diffusion model to supervise the optimization of Neural Radiance Fields (NeRF). Despite its foundational significance, the original SDS framework faces challenges such as low geometric resolution, texture over-saturation, and multi-view macro-structural inconsistencies (e.g., the ``Janus problem''). To address these limitations, subsequent research has primarily focused on extending and refining the SDS framework along the following three dimensions:
\paragraph{(1) Geometry-Texture Decoupling.}
To mitigate the high computational costs of NeRF under high-resolution optimization, Magic3D~\cite{lin2023magic3d} introduced a coarse-to-fine two-stage optimization strategy. It initiates by obtaining coarse geometry via low-resolution diffusion priors, then employs Deep Marching Tetrahedra (DMTet) to extract high-resolution meshes, and finally optimizes high-frequency textures through differentiable rendering in the second stage. Fantasia3D~\cite{chen2023fantasia3d} further fully disentangled geometry modeling from appearance generation, independently distilling them using geometry normal maps and appearance RGB maps, respectively, which significantly improved the surface fidelity of the generated results. TextMesh~\cite{tancik2023textmesh} proposed a novel representation based on the Signed Distance Function (SDF), aiming to generate meshes that are more consistent with topological structures. RichDreamer~\cite{qiu2024richdreamer} identified that pure RGB supervision is insufficient for generating fine geometry and trained a generalizable ``Normal-Depth Diffusion Model'' as a strong geometric prior, notably enhancing the richness of surface details. Addressing the generation of complex semantics, Qin et al.~\cite{qin2025apply} proposed the Hierarchical-Chain-of-Generation, leveraging Large Language Models (LLMs) to decompose complex descriptions and resolve semantic confusion in multi-attribute objects. Similarly, LEGO-Maker~\cite{zhang2025legomaker} further improved structural plausibility through a semantic-driven modular generation strategy. Hybrid priors also represent a significant direction; 3DTopia~\cite{hong20243dtopia} proposed a novel two-stage framework that first utilizes 3D Tri-plane diffusion priors for rapid coarse shape generation, followed by 2D priors for texture refinement. Similarly, Ding et al.~\cite{ding2024text} introduced a bidirectional diffusion framework that synergizes 2D and 3D priors to enhance geometric consistency. However, such methods involving multi-stage pipelines or additional priors often incur increased training costs, and their performance may exhibit degradation when handling prompts not covered by the 3D priors.Furthermore, approaches like Latent-NeRF~\cite{metzer2023latent} and VolumeDiffusion~\cite{tang2023volumediffusion} explore optimization within latent or volumetric spaces to enhance generation flexibility and efficiency.
\paragraph{(2) Optimization of Distillation Objectives.}
Targeting the phenomena of texture ``over-smoothing'' and color ``over-saturation'' caused by SDS, ProlificDreamer~\cite{wang2023prolificdreamer} achieved a breakthrough with Variational Score Distillation (VSD) by modeling 3D parameter distributions. CSD (Classifier Score Distillation)~\cite{wang2023score} identified that excessive weights of Classifier-Free Guidance (CFG) in SDS are the primary cause of artifacts and proposed a correction scheme based on classifier scores. ScaleDreamer~\cite{ma2024scaledreamer} focused on resolution challenges, achieving higher-resolution 3D generation via an Asynchronous Score Distillation (ASD) mechanism without increasing memory overhead. LucidDreamer~\cite{liang2024luciddreamer} introduced Interval Score Matching (ISM), utilizing deterministic diffusion trajectories to counteract the over-smoothing induced by stochastic gradients, thereby significantly enhancing generation fidelity. Addressing the trade-off between shape and texture caused by Negative Prompts in SDS, Target-Balanced Score Distillation (TBSD)~\cite{xu2025target} proposed a multi-objective optimization framework that achieves a balanced enhancement of texture realism and geometric accuracy through an adaptive balancing strategy. Additionally, DaCapo~\cite{huang2025dacapo} modeled score distillation as Stacked Diffusion Bridges, effectively mitigating gradient conflicts and inefficiencies in traditional optimization processes. To further address the issue of non-continuity between texture and geometry, researchers have explored various enhancement strategies.
\paragraph{(3) Enhancement of Multi-view Consistency.} 
Addressing the most challenging Janus (multi-face) problem, MVDream~\cite{shi2024mvdream} and SweetDreamer~\cite{li2024sweetdreamer} fine-tune diffusion models on large-scale 3D datasets. This equips them with the capability to output multi-view consistent images, effectively circumventing geometric conflicts arising from independent view optimization and guiding the generation of 3D structures with global spatial coherence. Another avenue involves introducing explicit regularization. For instance, PeRF~\cite{rossi2022perf} utilizes keypoint detection to constrain object pose, while GeNVS~\cite{chan2023generative} employs novel view synthesis techniques to assist distillation. Furthermore, image-based perceptual regularization has been widely adopted. DietNeRF~\cite{jain2021dietnerf} leverages high-level semantic features from CLIP to constrain semantic consistency in unseen views. Meanwhile, some works utilize LPIPS~\cite{zhang2018unreasonable} loss to constrain texture similarity between adjacent views, attempting to smooth view transitions within Euclidean space. CAT3D~\cite{gao2024cat3d} and its successor CAT4D~\cite{gao2025cat4d} propose utilizing Video Diffusion Models as priors. By simulating real-world camera paths to constrain 3D spatial consistency, they significantly reduce multi-face artifacts. On the other hand, to inject 3D awareness into existing diffusion models, 3D-Adapter~\cite{chen20243d} designed a plug-and-play geometric consistency module, which guides the model to generate geometrically consistent multi-view images via 3D feedback enhancement during denoising steps.
Although existing works have made strides in their respective focuses, current SDS variants remain constrained by two major bottlenecks when handling complex topologies and fine textures: the ``Lack of Macro-topological Constraint'' and ``Insufficient Micro-statistical Alignment.'' On one hand, even methods introducing multi-view priors (e.g., MVDream~\cite{shi2024mvdream}) lack explicit logical constraints on the global topology of 3D objects (such as ``front-back exclusivity''). When facing prompts that are abstract or uncovered by training data, models are susceptible to degradation, leading to persistent non-physical structural redundancies (i.e., the Janus problem). On the other hand, existing regularization metrics (e.g., LPIPS) typically operate based on Euclidean distance, making it difficult to capture the complex statistical regularities of texture features in the manifold space. When confronting gradient noise from diffusion models, these methods fail to effectively align at the statistical level, resulting in uneliminated micro-geometric misalignment and texture tearing during view transitions. Therefore, constructing a unified framework capable of simultaneously constraining macro-semantic topology and smoothing micro-statistical distributions is key to achieving high-fidelity text-to-3D generation.
%% ================= Section 3: Methodology =================
\section{Methodology}
\subsection{Overall Framework}
\begin{figure*}[ht]
  \centering
  \Description{MOC-3D overall framework diagram: SPD Manifold-based Feature constraints and semantic view-order constraints.}
  \includegraphics[width=\linewidth]{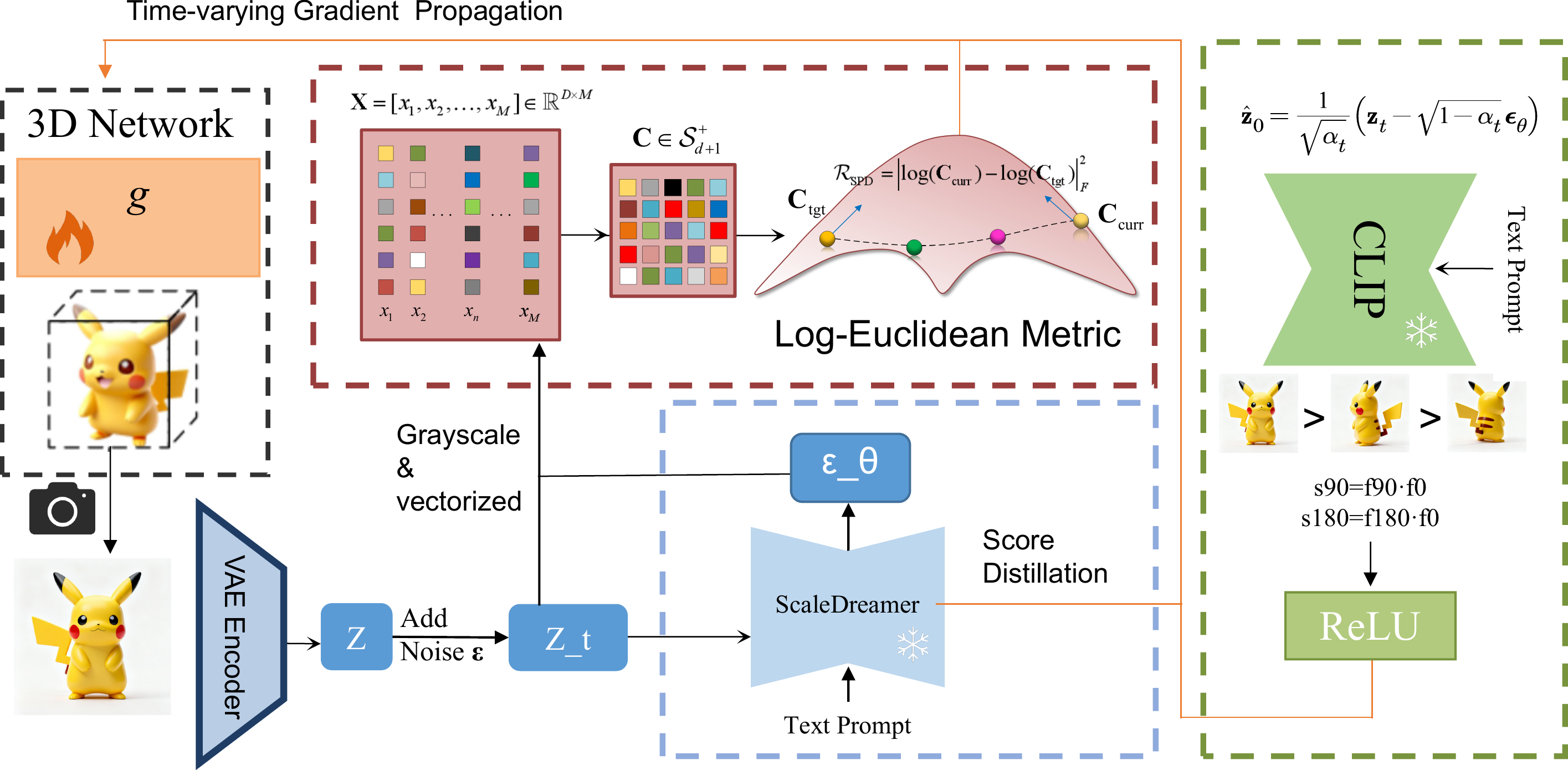}
  \caption{MOC-3D Collaborative Optimization Framework: Combining SPD Manifold Geometry Constraints and Semantic View-Order Constraints.}
  \label{fig:framework}
%%  \vspace{-5pt}%%fn
\end{figure*}

To address the macro-topological inconsistencies stemming from the view bias of 2D priors and the micro-geometric discontinuities arising from gradient noise, we propose MOC-3D, a novel optimization framework. This framework aims to resolve core challenges across two dimensions: inconsistency in macro-level topological structures and discontinuity in micro-level geometry. The overall network architecture is illustrated in Figure~\ref{fig:framework}.
The proposed MOC-3D framework incorporates two constraint modules designed to impose explicit constraints on the NeRF optimization process from the dimensions of semantic logic and manifold geometry, respectively:
\begin{itemize}
    \item \textbf{The Semantic View-Order Constraint Module} enforces a Monotonicity Rank Constraint based on CLIP scores. It provides global structural guidance to effectively rectify macro-topological inconsistencies.
    \item \textbf{The Manifold-based Feature Continuity Module} constructs a Riemannian Regularization Term based on the Symmetric Positive Definite (SPD) manifold. This addresses high-frequency artifacts and micro-geometric discontinuities at a statistical level.
\end{itemize}
During training, these two regularization terms are integrated with the baseline Score Distillation Sampling loss to form a unified global optimization objective. Gradient signals from this objective are jointly back-propagated to update the NeRF network parameters, ultimately guiding the model to converge towards high-quality 3D assets that feature both high-fidelity micro-details and correct macro-topology.
\subsection{Semantic View-Order Constraint Module}
Score Distillation based on pre-trained 2D diffusion models can provide rich texture information, but inherently lacks internal perception of global 3D topology. This absence of awareness causes the model to easily fall into local optima during optimization, leading to macro-topological inconsistencies, such as erroneously generating repetitive facial features on the back of an animal (i.e., the Janus problem). Such macro-level semantic ambiguities cannot be eliminated solely through local pixel-level optimization; explicit global semantic logic constraints must be introduced.
To bridge this gap in macro-cognitive capability, we introduce the Semantic View-Order Constraint Module. The core idea is to leverage the strong semantic priors of large-scale pre-trained Vision-Language Models (CLIP) to construct a ``Semantic View-Order Alignment Term.'' By explicitly imposing a semantically consistent Monotonicity Rank Constraint on semantic representations across different views, the module forces the model to learn the mapping relationship between view changes and semantic similarity, thereby constraining the global topological structure of the generated object.
Specifically, the computational flow of this module consists of two key steps: first, \textit{Semantic Feature Extraction}, which reconstructs images from the noisy latent space and maps them to a high-dimensional semantic space; and second, \textit{Monotonic Similarity Constraint}, which constructs a rank-based regularization term to penalize semantic distributions that violate view-order logic.
\subsubsection{Semantic Feature Extraction}
In each step of Score Distillation optimization, to obtain image representations usable for semantic evaluation, a noise-free approximation of the image must first be estimated from the current noisy latent variable $z_t$. Based on the diffusion model's scheduling coefficient $\alpha_t$, the denoised latent feature $\hat{z}_0$ can be estimated by:
\begin{equation}
    \hat{z}_0 = \frac{z_t - \sigma_t \epsilon_\phi(z_t, t, y)}{\alpha_t}
\end{equation}
where $\alpha_t$ denotes the scheduling coefficient for the forward noise addition process.
Subsequently, $\hat{z}_0$ is mapped back to the pixel space via the VAE decoder to obtain the reconstructed image $\hat{x}_0$. For a set of $N$ predicted clean images $\hat{\mathbf{x}}_0$ rendered from different azimuth angles, we utilize a pre-trained, frozen CLIP image encoder $\operatorname{CLIP}(\cdot)$ to extract their respective semantic feature vectors. These vectors are then L2-normalized to obtain a set of high-dimensional semantic embeddings $\{\mathbf{e}_{i} \in \mathbb{R}^{D_{\text{CLIP}}}\}_{i=1}^{N}$.
\subsubsection{Monotonic Similarity Constraint}
Studies indicate that the root cause of macro-semantic ambiguities like the Janus problem lies in the generative model's failure to learn the semantic decay patterns across views. To this end, we introduce a structured ranking regularization mechanism. This mechanism aims to establish a negative correlation between view deviation and semantic similarity. In specific implementation, the front view with an azimuth of $0^\circ$ is selected as the reference anchor, and its semantic embedding is denoted as $\mathbf{v}_{\text{ref}}$. For the remaining sequence of views sorted by increasing angle (e.g., $90^\circ, 180^\circ$), we calculate their cosine similarity with the reference view. For objects possessing inherent canonical orientation (e.g., characters, animals), the semantic similarity relative to the frontal text description typically exhibits a structured decay as the viewing angle deviates. We empirically model this as a \textbf{monotonic trend}: $S(0^\circ) > S(90^\circ) > S(180^\circ)$.
To ensure the model adheres to this \textbf{soft constraint}, we design a ranking-based Hinge Loss to penalize violations of the view-order prior. For a view sequence of length $K$, this regularization term is defined as:
\begin{equation}
\mathcal{R}_{\text{SVO}} = \sum_{i=1}^{K-1} \max \left( 0, \operatorname{sim}(v_{i+1}, v_{\text{ref}}) - \operatorname{sim}(v_{i}, v_{\text{ref}}) + \delta \right)
\end{equation}
where $\delta$ denotes the slack margin that enforces a minimum semantic distance between adjacent views. This margin ensures that the optimization receives persistent gradient signals until a clear semantic hierarchy is established, effectively preventing the "Janus problem" by explicitly separating the feature representations of distinct viewpoints.This objective function accurately captures and penalizes any violations of the geometric-semantic prior that ``farther views are less semantically relevant.'' The loss value converges to 0 if and only if the entire sequence satisfies the strict property of monotonic decrease. This structured prior provides explicit constraints for the model to learn reasonable spatial layouts and macro-topological relationships, thereby effectively resolving topological anomalies such as the multi-face effect.
From the perspective of optimization, the gradient of $\mathcal{R}_{\text{SVO}}$ with respect to the NeRF parameters remains non-zero only when the semantic similarity difference between $v_{i+1}$ and $v_{i}$ is less than the margin $\delta$. This selective back-propagation guides the generator to prioritize the correction of macro-topological reversals while avoiding unnecessary fine-tuning once the logical view-order is satisfied.
%% ================= 3.3 Manifold-based Feature Continuity Module =================
\subsection{Manifold-based Feature Continuity Module}
The Symmetric Positive Definite (SPD) manifold provides a Riemannian geometric framework for processing covariance matrices with specific geometric structures~\cite{pennec2006riemannian, arsigny2006log}. Unlike linear Euclidean space, the SPD manifold, through its non-linear Riemannian metric, effectively captures the second-order statistical information of feature sets (such as texture distribution density and spatial correlation) and inherently possesses invariance to affine transformations (e.g., translation and rotation). This characteristic has demonstrated strong robustness in computer vision, such as analyzing brain microstructures in medical imaging~\cite{basser1994mr} or handling complex illumination and pose variations in fine-grained classification~\cite{tuzel2006region}. Notably, works like SPDNet~\cite{huang2017riemannian} have further paved the way for integrating Riemannian geometry with deep neural networks, making direct feature learning in the manifold space feasible.
Despite the widespread application of SPD manifolds in discriminative tasks, their potential as a regularization mechanism in text-to-3D generation tasks remains underexplored. Existing generation methods predominantly rely on metrics in Euclidean space (e.g., MSE or LPIPS). While traditional style transfer utilizes Gram matrices in Euclidean space, we argue that mapping these second-order statistics to the Riemannian manifold provides a more natural geometry for texture alignment; complex visual textures (e.g., fur, fabric) are intrinsically defined by the statistical distribution of local features rather than isolated pixels. Forcing the alignment of these features with high-dimensional statistical regularities in Euclidean space often fails to capture their intrinsic laws, leading to feature-level discrepancies that manifest as micro-geometric discontinuities such as geometric misalignment and high-frequency artifacts.
Therefore, we introduce a geometric constraint mechanism based on the Symmetric Positive Definite (SPD) manifold. The core idea is to model the feature covariance matrices of multi-view rendered images as points on the SPD manifold and leverage the statistical alignment capability of Riemannian geometry to constrain the smooth evolution of feature second-order statistics. This approach elevates the metric dimension from ``pixel/feature-level alignment'' to ``statistical distribution-level alignment.'' It implements the decoupling of texture statistical properties from specific spatial positions, thereby granting the generator greater freedom to synthesize vivid high-frequency details and effectively eliminating discontinuities at view transitions.
Specifically, the computational workflow of this module comprises two key steps: first, \textit{SPD Matrix Construction and Feature Aggregation}, which encodes multi-view features into compact second-order statistical descriptors on the Riemannian manifold space; and second, \textit{Riemannian Metric-based Geometric Constraint}, which utilizes the Log-Euclidean Metric (LEM) to construct a regularization term, constraining the smooth evolution of micro-textures at a statistical level.
\subsubsection{SPD Matrix Construction and Feature Aggregation}
To comprehensively capture the global statistical properties of the object, we design a construction pipeline involving feature dimensionality reduction, statistical aggregation, and regularization.
Given a set of $N$ multi-view, four-channel latent feature maps rendered by NeRF, denoted as $\mathcal{F} = \{\mathbf{f}_i \in \mathbb{R}^{4 \times H \times W} \}_{i=1}^N$. Since directly calculating covariance matrices for high-dimensional channels is computationally expensive and prone to redundancy, we first execute luminance-based feature compression. By introducing a fixed weighting vector $\mathbf{w}$, we perform a weighted sum on each feature map, compressing it into a single-channel grayscale map $\mathbf{g}_i \in \mathbb{R}^{H \times W}$ representing luminance information. This operation significantly reduces the dimensionality for subsequent calculations while preserving the core texture structure.

Next, we perform global feature aggregation. The pixel values of all $N$ grayscale maps are flattened and concatenated to construct a feature vector set $\mathbf{X} \in \mathbb{R}^{D \times M}$ representing the global appearance distribution of the object (where $D$ represents the feature dimension and $M$ is the number of samples). Based on this set, we calculate its first-order statistics (mean vector $\boldsymbol{\mu}$) and second-order statistics (covariance matrix $\boldsymbol{\Sigma}$). To encode both mean and variance information within a unified mathematical object and ensure the matrix satisfies the symmetric positive definite property, we construct an extended SPD matrix $\mathbf{C}$. To ensure numerical stability and strict positive definiteness, the covariance matrix undergoes regularization: $\boldsymbol{\Sigma}_{\text{reg}} = \boldsymbol{\Sigma} + \epsilon \mathbf{I}$, where $\epsilon$ is a small positive constant. Subsequently, we construct a block matrix in the following form:
\begin{equation}
    \mathbf{C} = \begin{pmatrix}
    \boldsymbol{\Sigma}_{\text{reg}} + \boldsymbol{\mu}\boldsymbol{\mu}^{\top} & \boldsymbol{\mu} \\
    \boldsymbol{\mu}^{\top} & 1
    \end{pmatrix}
\end{equation}

% --- 下面是重点修改的段落，回应了 R1 和 R4 ---
In this matrix, the bottom-right element is a scalar $1$, aligning with the homogeneous representation of multivariate distributions. The top-left sub-block is essentially the second moment matrix about the origin, which implicitly fuses variance and mean information. Crucially, fusing the mean and covariance into a single extended SPD representation provides a distinct mathematical advantage over aligning them separately with simpler independent losses. This joint representation allows the subsequent Riemannian metrics to intrinsically capture the cross-correlations between first-order mean shifts and second-order variance scaling. This construction ensures that $\mathbf{C}$ always resides on the SPD manifold $\mathcal{S}_{D+1}^+$, laying a rigorous mathematical foundation for treating the feature distribution as a holistic geometric entity.
\subsubsection{Riemannian Metric-based Geometric Constraint}
Upon successfully constructing the SPD matrix, the next step is to define a distance metric on the manifold space to quantify the discrepancy between the current generation state and the target statistical distribution.
Within the academic community, various distance definitions exist on the SPD manifold, with the Affine Invariant Riemannian Metric (AIRM) and the Log-Euclidean Metric (LEM) being the most prominent~\cite{pennec2006riemannian}. Although AIRM possesses excellent theoretical properties, its calculation involves computationally expensive matrix inversions and iterative processes, making it difficult to meet the efficiency demands of deep neural network training. In contrast, LEM utilizes matrix logarithm operations to map the manifold of SPD matrices $\mathcal{S}_{d+1}^+$ to its associated Tangent Space $\mathcal{T}_{\mathbf{I}}\mathcal{S}_{d+1}^+$ (the Lie algebra of symmetric matrices), which forms a linear vector space. This allows us to transform complex geodesic distance calculations into Euclidean distance calculations within the tangent space, preserving the transformation invariance of Riemannian geometry while achieving high computational efficiency and numerical stability. Therefore, we adopt LEM to quantify the difference between the current generation state $\mathbf{C}_{\text{curr}}$ and the target statistical distribution $\mathbf{C}_{\text{tgt}}$. Specifically, both matrices are constructed using the multi-view aggregation pipeline defined in Equation (3). While $\mathbf{C}_{\text{curr}}$ is calculated from the multi-view latent features rendered at the current optimization step, the target matrix $\mathbf{C}_{\text{tgt}}$ is derived identically from the feature distributions of the target reference images. Formally, $\mathcal{R}_{\text{SPD}}$ is defined as the squared Frobenius Norm of the difference between the two SPD matrices in the log-domain:
% --- 替换后的新句如上 ---
% --- 替换前的原句 ---
% Therefore, we adopt LEM to quantify the difference between the feature covariance matrix of the current rendered view $\mathbf{C}_{\text{curr}}$ and the target statistical distribution $\mathbf{C}_{\text{tgt}}$. Formally, $\mathcal{R}_{\text{SPD}}$ is defined as the squared Frobenius Norm...

%%
%%\begin{equation}
%%    \mathcal{R}_{\text{SPD}} = \left\| \log(\mathbf{C}_{\text{curr}}) - \log%%(\mathbf{C}_{\text{tgt}}) \right\|_F^2
%%\end{equation}
%%
\begin{equation}
    \mathcal{R}_{\text{SPD}} = \left\| \log(\mathbf{C}_{\text{curr}} + \epsilon \mathbf{I}) - \log(\mathbf{C}_{\text{tgt}} + \epsilon \mathbf{I}) \right\|_F^2
\end{equation}
where $\log(\cdot)$ denotes the matrix logarithm operator, and $\epsilon=10^{-6}$ is a Tikhonov regularization term ensuring numerical stability and strict positive definiteness during back-propagation. This constraint enforces the model to minimize statistical distribution discrepancies within the Riemannian manifold. Unlike Euclidean-based losses that often impose overly restrictive pixel-wise correspondences, our SPD-based regularization focuses on aligning second-order statistics (e.g., feature correlations and directional orientations). Due to its intrinsic spatial invariance, this manifold constraint allows the generator to synthesize diverse high-frequency details while maintaining a globally consistent statistical structure. Consequently, minimizing this loss effectively suppresses view-dependent artifacts and structural discontinuities, fostering smooth and natural texture transitions across complex geometries.
In summary, by minimizing the proposed Riemannian geometric regularization term, the MOC-3D framework achieves explicit optimization of the visual high-dimensional statistical properties of the generated results directly within the Riemannian manifold space.
%% ================= 3.4 Dynamic Optimization Strategy =================
\subsection{Dynamic Optimization Strategy based on Time-Varying Constraint Scheduling}
To achieve multi-dimensional synergistic optimization within a single end-to-end training pipeline, we construct a dynamic joint optimization objective function. By integrating the baseline generation loss with the proposed geometric and semantic constraints via weighted aggregation, it is defined as follows:
\begin{equation}
    \mathcal{L}_{\text{total}}^{(t)} = \mathcal{L}_{\text{ASD}} + \lambda_{\text{SPD}}(t)\mathcal{R}_{\text{SPD}} + \lambda_{\text{SVO}}(t)\mathcal{R}_{\text{SVO}}
\end{equation}
where $\mathcal{L}_{\text{ASD}}$ is the baseline Asynchronous Score Distillation (ASD) loss providing basic generation capabilities, while $\lambda_{\text{SVO}}(t)$ and $\lambda_{\text{SPD}}(t)$ are time-varying weighting coefficients. Our core optimization logic follows the structural evolution laws of 3D generation. In the early training stage ($t < T_{\text{warmup}}$), the model is in the topology establishment phase, where $\lambda_{\text{SVO}}$ is assigned a high initial value. This aims to leverage semantic view-order priors to forcibly rectify spatial layout ambiguities, laying a solid macro-structural foundation for the model. As training progresses, $\lambda_{\text{SVO}}$ follows a linear annealing strategy to gradually decay, unleashing the model's freedom to express high-frequency details. Conversely, the Manifold-based Feature Continuity Constraint ($\lambda_{\text{SPD}}$) maintains a high activation level throughout the cycle and further consolidates its dominant position in the mid-to-late stages. This design aims to continuously suppress geometric artifacts introduced by the stochasticity of score sampling via smooth constraints on the Riemannian manifold, ensuring that generated entity surfaces conform to physically intuitive second-order continuity while possessing high-fidelity texture mapping. This coarse-to-fine dynamic scheduling mechanism effectively alleviates gradient conflict issues during synergistic training, ultimately guiding the model to converge to the global optimal solution.
% === 4.实验部分 ===
\section{Experiments}
\subsection{Experimental Design}
\paragraph{Evaluation Benchmark.} 
Consistent with the zero-shot generation paradigm, our experiments do not rely on a specific 3D training set but are evaluated using a diverse collection of text prompts covering various topological complexities and texture characteristics. This benchmark primarily comprises standard test cases from prior works such as DreamFusion~\cite{poole2023dreamfusion}, assessing generation capabilities in general scenarios. Furthermore, to rigorously evaluate the model's performance in fine-grained generation tasks and its ability to handle dual consistency challenges, we extend this dataset with a specific set of prompts featuring distinct cultural styles, complex geometric details, and high-frequency repetitive patterns (e.g., Jiangnan-style traditional dwellings).
\paragraph{Implementation Details.} 
All training and inference experiments are conducted on a single NVIDIA RTX A6000 GPU. Our MOC-3D framework is built upon the open-source ScaleDreamer~\cite{ma2024scaledreamer} codebase and utilizes the Adam optimizer for parameter updates. Regarding the underlying 3D representation, we adopt the multi-resolution Hash Grid from Instant-NGP~\cite{muller2022instant} combined with VolSDF~\cite{yariv2021volume} for volume rendering. This hybrid representation supports high-fidelity surface geometry reconstruction inherent to SDF-based modeling, providing a robust continuous geometric foundation for subsequent micro-texture optimization, while the Hash Grid acceleration mitigates the computational overhead. To ensure reproducibility and fair comparison, a fixed random seed is set for all test cases, and input text formats are standardized.
\paragraph{Evaluation Metrics.} 
We employ a comprehensive evaluation protocol combining quantitative analysis and qualitative assessment to holistically measure generation quality.
\paragraph{Quantitative Analysis.} For each sample, we uniformly sample and render 120 images from surrounding views. We utilize three key metrics. \textbf{CLIP R-Precision (R@1)} measures the semantic consistency between generated content and input prompts by calculating the retrieval accuracy of rendered images among candidate texts. \textbf{CLIP Score (Sim)} calculates the cosine similarity between the features of rendered images and input texts, reflecting the degree of text-image semantic alignment. \textbf{View-Consistency LPIPS} specifically quantifies micro-geometric discontinuity by calculating the perceptual distance between adjacent views (with a fixed interval). A lower value indicates smoother texture transitions, implying fewer geometric misalignments.
\paragraph{Qualitative Evaluation.} We render 360-degree videos and multi-view close-up images for visual inspection. The assessment focuses on two dimensions: \textbf{Macro-topological Logic} (checking for non-physical structural redundancies like the Janus problem) and \textbf{Micro-surface Continuity} (checking for high-frequency texture artifacts and local geometric dislocations).
\begin{comment}
\paragraph{Experimental Design Overview.} 
The experiments are structured into three core components: 
(1) \textit{Comparative Analysis} against mainstream methods, which encompasses qualitative visual inspection, quantitative metric evaluation, and subjective user studies; 
(2) \textit{Ablation Study} to validate the individual contributions of the proposed modules; 
and (3) \textit{Generalizability Experiments} to assess performance in specific cultural heritage scenarios.
\end{comment}
\subsection{Comparative Analysis against Mainstream Methods}
To comprehensively evaluate MOC-3D, we conduct a cross-comparison against four representative baselines: \textbf{DreamFusion}~\cite{poole2023dreamfusion} (pioneering SDS baseline), \textbf{ScaleDreamer}~\cite{ma2024scaledreamer} (our backbone framework), \textbf{GaussianDreamer}~\cite{chen2023text} (benchmark for micro-texture), and \textbf{Hunyuan3D}~\cite{yang2024Hunyuan3D} (industrial state-of-the-art). Experiments are performed under unified conditions, utilizing open-source implementations~\cite{guo2023threestudio} for DreamFusion and GaussianDreamer, and the official service for Hunyuan3D.
%% === Figure 2: Qualitative Comparison ===
\begin{figure}[t]
    \centering
    \Description{Qualitative comparison with mainstream methods: MOC-3D vs DreamFusion, ScaleDreamer, GaussianDreamer, Hunyuan3D.}
    \includegraphics[width=\linewidth]{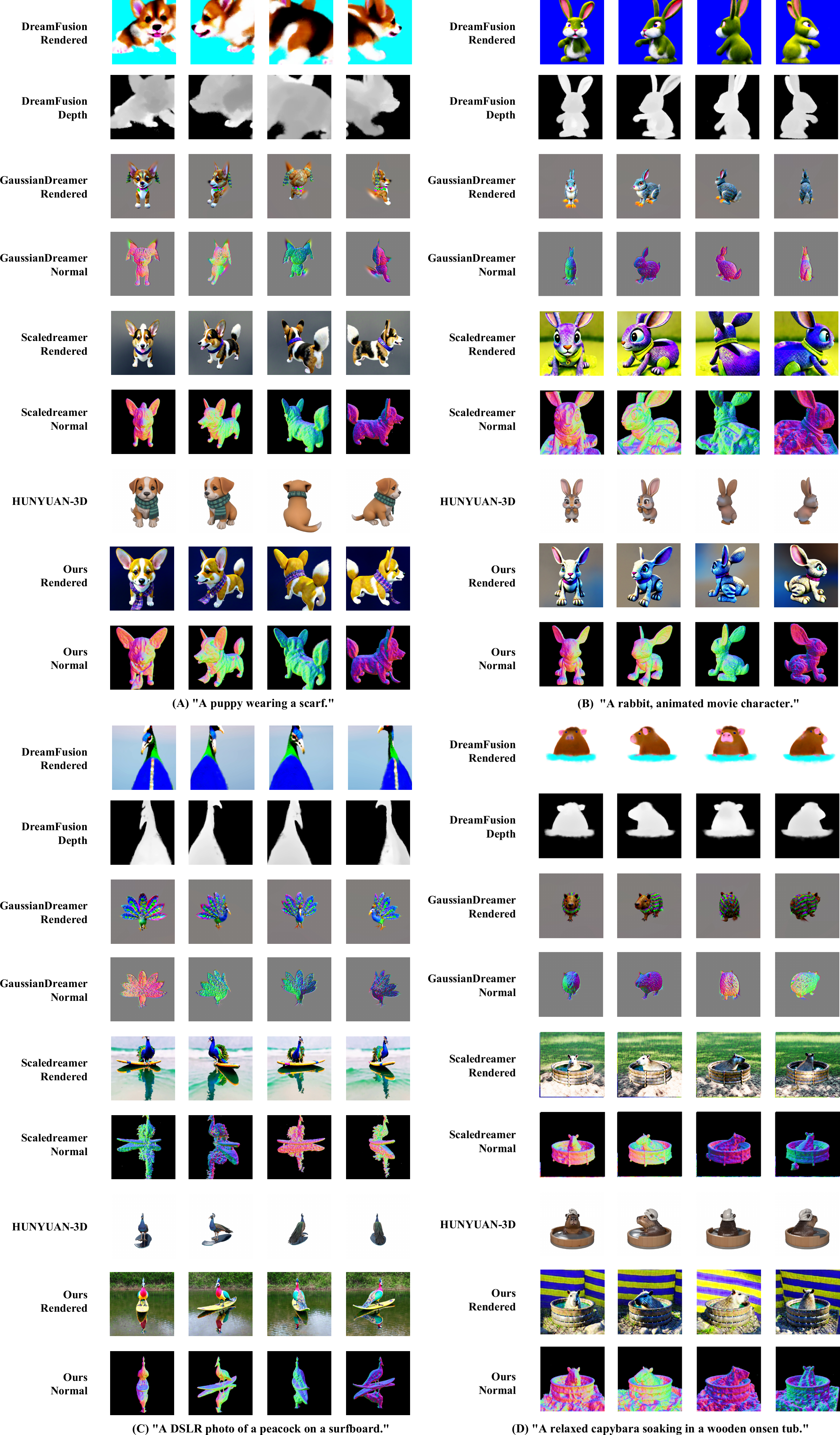} 
    \caption{Qualitative comparison with mainstream methods. Our MOC-3D generates high-fidelity 3D assets with consistent macro-topology and continuous micro-textures.}
    \label{fig:qual_comparison}
    %%\vspace{-5pt}%fn
\end{figure}
\paragraph{Qualitative Evaluation.}
Figure~\ref{fig:qual_comparison} presents the visual comparisons. Baselines frequently exhibit significant failures in two dimensions. 
Regarding \textbf{macro-topological consistency}, DreamFusion and ScaleDreamer suffer from non-physical Janus problems (e.g., the two-headed peacock in \textbf{Row C} and multi-faced capybara in \textbf{Row D}) or structural incompleteness. 
Regarding \textbf{micro-surface continuity}, ScaleDreamer displays severe high-frequency noise and geometric discontinuities (e.g., the rabbit fur in \textbf{Row B}), while Hunyuan3D tends to over-smooth textures, losing realistic details (e.g., the dog fur in \textbf{Row A}). 
In contrast, MOC-3D successfully constructs stylistically accurate characters with semantically consistent spatial relationships and smooth, high-fidelity surfaces, effectively overcoming the inherent view bias of standard SDS.
\paragraph{Quantitative Evaluation and User Study.}
%% === 表 1 ===
\begin{table}[t]
    \centering
    \Description{Table comparing MOC-3D to DreamFusion, ScaleDreamer, and HunYuan-3D on CLIP similarity, R@1 retrieval, and view-consistency LPIPS.}
    \caption{Quantitative comparison with mainstream methods. We report CLIP Similarity (Sim), R-Precision (R@1), and LPIPS. $\uparrow$ indicates higher is better, $\downarrow$ indicates lower is better.}
    \label{tab:quant_comparison}
    \begin{tabular}{l|c|c|c}
      \toprule
      Method & Sim $\uparrow$ & R@1 $\uparrow$ & LPIPS $\downarrow$ \\
      \midrule
      DreamFusion \cite{poole2023dreamfusion} & 0.231 & 0.88 & 0.351 \\
      ScaleDreamer \cite{ma2024scaledreamer} & 0.245 & 0.94 & 0.301 \\
      HunYuan-3D \cite{yang2024Hunyuan3D} & \textbf{0.264} & 0.98 & \textbf{0.232} \\
      \midrule
      \textbf{MOC-3D (Ours)} & 0.263 & \textbf{1.00} & 0.237 \\
      \bottomrule
    \end{tabular}
  \end{table}
%% === 表 2 ===
\begin{table}[h]
    \centering
    \Description{User study mean opinion scores (1--5) for 3D structural consistency, texture quality, and prompt alignment across methods.}
    \caption{User Study Results (MOS $\uparrow$). We evaluate 3D Structural Consistency, Texture/Detail Quality, and Prompt Consistency on a scale of 1-5.}
    \label{tab:user_study}
    \begin{tabular}{l|c|c|c}
      \toprule
      Method & Structure & Texture & Alignment \\
      \midrule
      GaussianDreamer \cite{chen2023text} & 3.20 & 3.20 & 3.90 \\
      ScaleDreamer \cite{ma2024scaledreamer} & 3.53 & 3.77 & \textbf{4.30} \\
      HunYuan-3D \cite{yang2024Hunyuan3D} & 3.80 & 3.85 & 4.15 \\
      \midrule
      \textbf{MOC-3D (Ours)} & \textbf{4.00} & \textbf{3.95} & \textbf{4.30} \\
      \bottomrule
    \end{tabular}
  \end{table}
Table~\ref{tab:quant_comparison} summarizes the automated metrics. 
MOC-3D demonstrates strong semantic consistency (Sim/R@1), matching or exceeding high-performing baselines. 
Notably, in terms of perceptual quality, our method achieves a competitive View-Consistency LPIPS (0.237), comparable to state-of-the-art industrial frameworks (e.g., Hunyuan3D). 
This indicates that our approach effectively minimizes micro-discontinuities without relying on large-scale 3D training data.
Crucially, these objective results are reinforced by the User Study (Table~\ref{tab:user_study}, $N=30$), where volunteers blindly rated MOC-3D highest in \textbf{3D Structural Consistency} (\textbf{4.00}, surpassing ScaleDreamer's 3.53 and Hunyuan3D's 3.80) and \textbf{Texture Quality} (\textbf{3.95}). 
The alignment between competitive numerical metrics and superior human perceptual ratings validates our synergistic optimization: the Manifold-based Feature Continuity Constraint ($\mathcal{R}_{\text{SPD}}$) suppresses gradient noise to enhance texture fidelity, while the Semantic View-Order Constraint ($\mathcal{R}_{\text{SVO}}$) strictly enforces global topological correctness.
%% ================= 4.3 Ablation Study =================
\subsection{Ablation Study}
\begin{figure}[t]
  \centering
  \Description{Ablation study: visual comparison of baseline, geometric constraint only, semantic constraint only, and full MOC-3D (peacock and capybara cases).}
  \includegraphics[width=\linewidth]{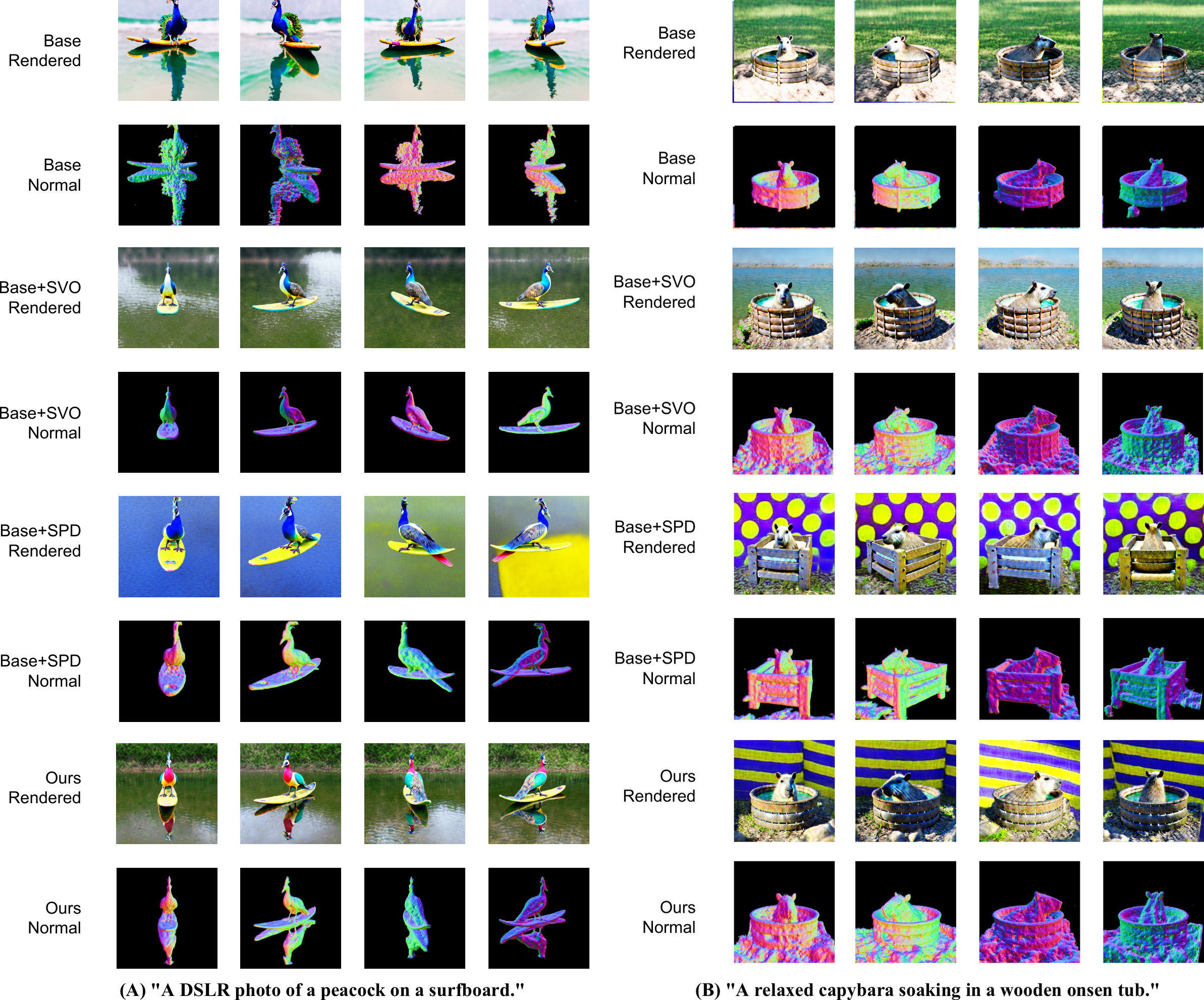}
  \caption{Ablation Study: Visual comparison of module contributions.}
  \label{fig:ablation_vis}
  %%\vspace{-5pt}%fn
\end{figure}

To systematically validate the independent contributions of each module within MOC-3D, we conducted an ablation study using the ScaleDreamer~\cite{ma2024scaledreamer} baseline. Visual comparisons are presented in Figure~\ref{fig:ablation_vis}, and quantitative metrics are summarized in Table~\ref{tab:ablation_quant}.
%% --- Fig 3: Ablation Study Visuals ---
\paragraph{Visual Analysis.}
The qualitative results in Figure~\ref{fig:ablation_vis} intuitively demonstrate the distinct mechanisms of each module. 
In the \textbf{``Peacock''} case (Fig.~3A), the \textbf{Baseline} exhibits limitations in dual dimensions: blurred feather textures (micro-discontinuity) and non-physical topological inconsistencies where facial features appear on the back. 
Introducing the \textbf{Manifold-based Feature Continuity Constraint ($\mathcal{R}_{\text{SPD}}$)} successfully sharpens the feather edges and smooths the surfboard surface; however, it fails to eliminate the multi-face effect, leaving the head orientation chaotic. 
Conversely, employing the \textbf{Semantic View-Order Constraint ($\mathcal{R}_{\text{SVO}}$)} rectifies the global topology to a single head orientation, yet results in over-smoothed surface details due to the lack of micro-statistical alignment. 
The \textbf{Full Model (Ours)} integrates both strengths, preserving rich, natural high-frequency details while ensuring a semantically consistent structure. 
This pattern is similarly reflected in the \textbf{``Capybara''} case (Fig.~3B). 
Where the Baseline suffers from severe gradient noise (messy fur), the Full Model vividly restores the fine qualities of wet fur and lighting details while maintaining correct spatial relationships, achieving an optimal balance between structure and texture.
%% --- Table 3: Ablation Quantitative ---
\begin{table}[h]
    \centering
    \Description{Ablation table reporting CLIP similarity, R@1, and LPIPS for baseline, adding semantic view-order loss, adding SPD manifold loss, and full MOC-3D.}
    \caption{Quantitative Results of the Ablation Study. We verify the contribution of $\mathcal{L}_{\text{SVO}}$ and $\mathcal{L}_{\text{SPD}}$.}
    \label{tab:ablation_quant}
    \begin{tabular}{l|c|c|c}
      \toprule
      Settings & Sim $\uparrow$ & R@1 $\uparrow$ & LPIPS $\downarrow$ \\
      \midrule
      Baseline & 0.231 & 1.000 & 0.3513 \\
      Base + $\mathcal{L}_{\text{SVO}}$ & 0.251 & 1.000 & 0.2619 \\
      Base + $\mathcal{L}_{\text{SPD}}$ & 0.227 & 1.000 & 0.2441 \\
      \midrule
      \textbf{MOC-3D (Ours)} & \textbf{0.263} & \textbf{1.000} & \textbf{0.2373} \\
      \bottomrule
    \end{tabular}
  \end{table}
\paragraph{Quantitative Verification.}
Table~\ref{tab:ablation_quant} further corroborates these visual findings with numerical evidence. 
Regarding micro-texture, the introduction of $\mathcal{R}_{\text{SPD}}$ leads to a significant reduction in the \textbf{LPIPS} metric compared to the Baseline, confirming its core role in suppressing high-frequency artifacts and smoothing feature distributions via Riemannian regularization. 
 Regarding macro-topology, while $\mathcal{R}_{\text{SVO}}$ serves to anchor structural logic (effectively resolving Janus problems), its solitary application yields a higher LPIPS, reflecting the texture over-smoothing observed qualitatively. 
Crucially, the \textbf{Synergy} of both modules enables the Full Model to achieve the best performance across all metrics, recording the highest CLIP Similarity (\textbf{0.263}) and the lowest LPIPS (\textbf{0.237}). 
This confirms that integrating macro-semantic anchoring with micro-statistical refinement is essential for high-fidelity generation.
\paragraph{Computational Cost Analysis.}
While the proposed MOC-3D framework significantly enhances generation quality, it introduces additional computational overhead. Evaluated on a single RTX A6000 GPU, the baseline ScaleDreamer achieves an optimization speed of approximately 1.33 iterations per second (it/s), whereas our full framework operates at around 0.46 it/s. This decrease in training efficiency is primarily attributed to two computationally intensive operations: the multi-view feature extraction via the CLIP text-image encoder in the SVO module, and the matrix logarithm calculations (which involve eigenvalue decompositions) for high-dimensional tensors in the SPD manifold constraint. We view this as an acceptable trade-off at the current stage, as exchanging a certain degree of computational efficiency for substantial improvements in macro-topological consistency and micro-texture fidelity aligns with the primary goal of generating high-quality 3D assets. Furthermore, as discussed in Section 5, we plan to address this limitation in future work by migrating our manifold constraints to more computationally efficient explicit representations like 3D Gaussian Splatting (3DGS).
%% ================= Section 5: Conclusion and Future Work =================
\section{Conclusion and Future Work}
In this paper, we propose MOC-3D, a novel synergistic optimization framework addressing the issues of macro-topological inconsistency and micro-geometric discontinuity in existing text-to-3D generation methods. Built upon ScaleDreamer, this framework innovatively introduces dual consistency constraints. 
At the \textbf{macro level}, the \textit{Semantic View-Order Constraint Module} leverages CLIP priors to construct a monotonicity rank constraint. This injects explicit global structural guidance into the optimization process, effectively rectifying multi-head topological anomalies such as the Janus problem. 
At the \textbf{micro level}, the \textit{Manifold-based Feature Continuity Module} models multi-view features as points on the Symmetric Positive Definite (SPD) manifold. Utilizing Riemannian geometric metrics, it constrains the smooth evolution of feature distributions at a statistical level, effectively suppressing high-frequency artifacts and geometric discontinuities induced by gradient noise.
To comprehensively validate the effectiveness of our method, we designed a systematic experimental protocol and drew the following conclusions. 
First, on test prompts covering diverse topological and textural characteristics, MOC-3D outperforms mainstream baselines (e.g., DreamFusion, ScaleDreamer, and Hunyuan3D) in key metrics such as Semantic Consistency (CLIP Score) and Perceptual Quality (LPIPS), effectively resolving multi-face effects and texture inconsistencies. 
Second, ablation studies confirm the effectiveness of the dual constraints, indicating that the Semantic View-Order Constraint effectively rectifies macro-topological errors, while the Manifold-based Feature Continuity Constraint is responsible for smoothing micro-texture details. Their synergy achieves an optimal balance between structure and texture. 
Third, in extended experiments on specific cultural style scenarios, MOC-3D demonstrates exceptional robustness. It successfully overcomes the limitations of baseline models in handling high-frequency repetitive textures and hybrid geometric structures, effectively avoiding texture over-smoothing or structural proportion misalignment caused by single-module constraints, thereby realizing high-fidelity generation of complex hybrid topologies.
\paragraph{Limitations and Future Work.} 
Although our method achieves high-quality text-to-3D generation and effectively addresses the dual consistency challenges, it has certain limitations. First, the Semantic View-Order Constraint relies on a heuristic monotonicity assumption ($S(0^\circ) > S(90^\circ) > S(180^\circ)$). While highly effective for objects with distinct canonical orientations (e.g., animals, vehicles), this soft constraint may introduce bias or become less informative when generating strictly symmetric objects (e.g., a uniform sphere) or abstract shapes lacking a defined front-back relationship. Future work could introduce category-aware or adaptive scheduling to dynamically bypass this constraint for symmetric prompts. Second, evaluating true 3D structural quality remains challenging. Due to the lack of ground-truth 3D assets in zero-shot text-to-3D tasks, our evaluation primarily relies on 2D rendering metrics (e.g., LPIPS, CLIP Score). Exploring robust, reference-free 3D geometric metrics is an important future direction.
Lastly, there remains potential for improvement in rendering and optimization speed. Future work will explore migrating the manifold statistical alignment mechanism to explicit representations such as 3D Gaussian Splatting (3DGS). By leveraging the discrete nature and efficient rasterization capabilities of 3DGS, we aim to achieve real-time training and inference for 3D object generation while maintaining high-fidelity texture constraints.
%%致谢
\begin{acks}
This work was supported by the Chongqing Social Science Planning Project (2025PY15), the Chongqing Municipal Education Commission Humanities and Social Sciences General Project (24SKGH077), the Chongqing Natural Science Foundation General Project \linebreak[4]
(CSTB2025NSCQ-GPX1037, CSTB2023NSCQ-MSX0407, CSTB2024NSCQ-MSX0468), the Science and Technology Research Project of Chongqing Education Commission (KJQN202300533), and the Doctoral Research Project of Chongqing Normal University (22XLB018, 22XLB017).
\end{acks}

%% ====================================================
%% ================= 参考文献 =================
%%  sample-base.bib 文件其中放入了 BibTeX 格式的参考文献
\clearpage %% 1. 强制换页，并输出所有缓存的图表
\balance%% 2. 参考文献的第一页 也双栏对齐，用 \balance
%%\nocite{*} %%强制显示所有参考文献
\bibliographystyle{ACM-Reference-Format}
\bibliography{sample-base}
\end{document}